\pdfoutput=1
\PassOptionsToPackage{table}{xcolor}
\documentclass[11pt]{article}

\usepackage[]{acl}

\usepackage{times}
\usepackage{latexsym}

\usepackage[T1]{fontenc}
\usepackage[utf8]{inputenc}

\usepackage{microtype}

\usepackage{inconsolata}

\usepackage{forest}
\usetikzlibrary{shadows}
\usepackage{amsmath}
\usepackage{amssymb}
\usepackage{cleveref}

\usepackage{tabularray}
\usepackage{booktabs}
\usepackage{siunitx}
\usepackage{etoolbox}
\usepackage{nicematrix}
\usepackage{multirow}
\usepackage{makecell}
\usepackage{enumitem}
\usepackage{xcolor,pifont}
\usepackage{fontawesome5}
\usepackage{hyperref}
\usepackage{subcaption}

\usepackage{amstext} % for \text macro
\usepackage{array}   % for \newcolumntype macro
\usepackage{arydshln}

\newcolumntype{L}{>{$}l<{$}} % math-mode version of "l" column type
\newcolumntype{R}{>{$}r<{$}}
\newcolumntype{C}{>{$}c<{$}}

\definecolor{lightcoral}{rgb}{0.94, 0.5, 0.5}
\definecolor{lightgreen}{rgb}{0.56, 0.93, 0.56}
\definecolor{cadmiumgreen}{rgb}{0.0, 0.42, 0.24}
\definecolor{dartmouthgreen}{rgb}{0.05, 0.5, 0.06}
\definecolor{ao}{rgb}{0.0, 0.5, 0.0}
\definecolor{alizarin}{rgb}{0.82, 0.1, 0.26}
\definecolor{harvestgold}{rgb}{0.85, 0.57, 0.0}
\definecolor{brightlavender}{rgb}{0.75, 0.58, 0.89}
\definecolor{capri}{rgb}{0.0, 0.75, 1.0}
\definecolor{carminepink}{rgb}{0.92, 0.3, 0.26}
\definecolor{celadon}{rgb}{0.67, 0.88, 0.69}
\definecolor{darkpastelgreen}{rgb}{0.01, 0.75, 0.24}
\definecolor{lightskyblue}{rgb}{0.53, 0.81, 0.98}
\definecolor{bisque}{rgb}{1.0, 0.89, 0.77}
\definecolor{lightpink}{rgb}{1.0, 0.71, 0.76}

\newcommand*\circled[1]{\tikz[baseline=(char.base)]{
            \node[shape=circle,draw,inner sep=0.5pt,thick] (char) {#1};}}

\crefformat{section}{\S#2#1#3} 
\crefformat{subsection}{\S#2#1#3}
\crefformat{subsubsection}{\S#2#1#3}

\newrobustcmd{\B}{\bfseries}
\newcommand*\colourcheck[1]{%
  \expandafter\newcommand\csname #1check\endcsname{\textcolor{#1}{\ding{52}}}%
}
\newcommand*\colourcross[1]{%
  \expandafter\newcommand\csname #1cross\endcsname{\textcolor{#1}{\ding{55}}}%
}

\newcommand{\R}[1]{\textcolor{alizarin}{#1}}
\newcommand{\G}[1]{\textcolor{dartmouthgreen}{#1}}

\newcommand{\eg}{\emph{e.g.}}
\newcommand{\ie}{\emph{i.e.}}

\colourcheck{blue}
\colourcheck{green}
\colourcheck{darkpastelgreen}
\colourcross{blue}
\colourcross{applegreen}
\colourcross{red}

\title{RetrievalQA: Assessing Adaptive Retrieval-Augmented Generation for Short-form Open-Domain Question Answering}
\author{Zihan Zhang\textsuperscript{1},  Meng Fang\textsuperscript{2}, Ling Chen\textsuperscript{1} \\
        \textsuperscript{1}University of Technology Sydney \
        \textsuperscript{2}University of Liverpool \\
        \texttt{Zihan.Zhang-5@student.uts.edu.au}, \texttt{Meng.Fang@liverpool.ac.uk} \\
        \texttt{Ling.Chen@uts.edu.au}
        }

\begin{document}
\maketitle
\begin{abstract}

% Retrieval-augmented generation (RAG) that augments large language models (LLMs) with the retrieval of relevant knowledge has become increasingly popular in knowledge-intensive tasks, including open-domain question-answering (QA). RAG improves task performance by reducing LLMs' hallucinations due to outdated parametric knowledge.

% Retrieval-augmented generation (RAG) reduces large language models' (LLMs) factual hallucinations by augmenting the input with retrieval of relevant knowledge, which has become increasingly popular in knowledge-intensive tasks, including open-domain question-answering (QA).
% However, standard RAG retrieves \textit{indiscriminately} regardless of the input query, which may result in suboptimal task performance and increased inference costs.

Adaptive retrieval-augmented generation (ARAG) aims to dynamically determine the necessity of retrieval for queries instead of retrieving indiscriminately to enhance the efficiency and relevance of the sourced information. However, previous works largely overlook the evaluation of ARAG approaches, leading to their effectiveness being understudied.
This work presents a benchmark, \texttt{RetrievalQA}, comprising 1,271 short-form questions covering new world and long-tail knowledge. 
% The knowledge required to answer the questions is absent from LLMs, making it a suitable testbed to evaluate existing ARAG methods.
The knowledge necessary to answer the questions is absent from LLMs; therefore, external information must be retrieved to answer correctly. This makes \texttt{RetrievalQA} a suitable testbed to evaluate existing ARAG methods.
We observe that calibration-based methods heavily rely on threshold tuning, while vanilla prompting is inadequate for guiding LLMs to make reliable retrieval decisions.
Based on our findings, we propose \textbf{T}ime-\textbf{A}ware \textbf{A}daptive \textbf{RE}trieval (\textbf{TA-ARE}), a simple yet effective method that helps LLMs assess the necessity of retrieval without calibration or additional training\footnote{The dataset and code are available at \url{https://github.com/hyintell/RetrievalQA}}.

\end{abstract}

\section{Introduction}
\label{sec_introduction}

Retrieval-augmented generation (RAG)  
 % \citep{guu2020retrieval, lewis2020retrieval, izacard2022atlas, ram-etal-2023-context}
  \citep{guu2020retrieval, lewis2020retrieval, ram-etal-2023-context}
that augments large language models (LLMs) with retrieval of relevant information has become increasingly popular in knowledge-intensive tasks, including open-domain question-answering (QA) \citep{zhang-etal-2023-large, kasai2023realtime, cui2023chatlaw,zhang-etal-2023-survey-efficient}.
However, standard RAG methods conduct retrieval \textit{indiscriminately}, irrespective of the input query, which may result in suboptimal task performance and increased inference costs \citep{gao2024retrievalaugmented}.
On one hand, LLMs encode vast knowledge in parameters through large-scale pre-training, enabling them to effortlessly handle straightforward
% \footnote{
% Here, "straightforward" refers to the questions that LLMs can easily answer correctly using their parametric knowledge.
% } 
queries without retrieval \citep{mallen-etal-2023-trust}. On the other hand, the retrieved context may contain noise and irrelevant information, and augmenting noisy context can potentially distract LLMs, thereby impeding task performance 
% \citep{pmlr-v202-shi23a, yoran2023making}.
\citep{pmlr-v202-shi23a}.

\begin{figure}[!t]
\centering
\begin{tabular}{c}
\begin{subfigure}{\columnwidth}
\centering
    \includegraphics[width=\columnwidth]{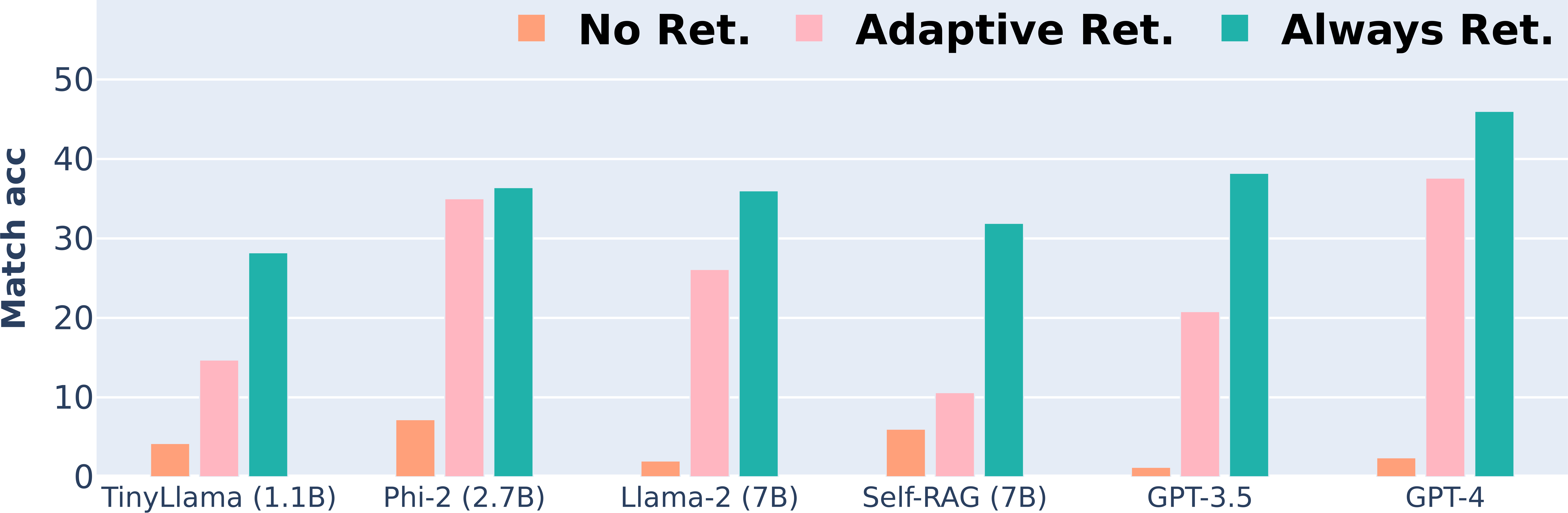}
    % \caption{
    % \footnotesize {No/Adaptive/Always Retrieval on \texttt{RetrievalQA}}
    % }
\end{subfigure}

\\
\hdashline
\\
% \medskip

\begin{subfigure}{\columnwidth}
\centering
    \includegraphics[width=0.7\columnwidth]{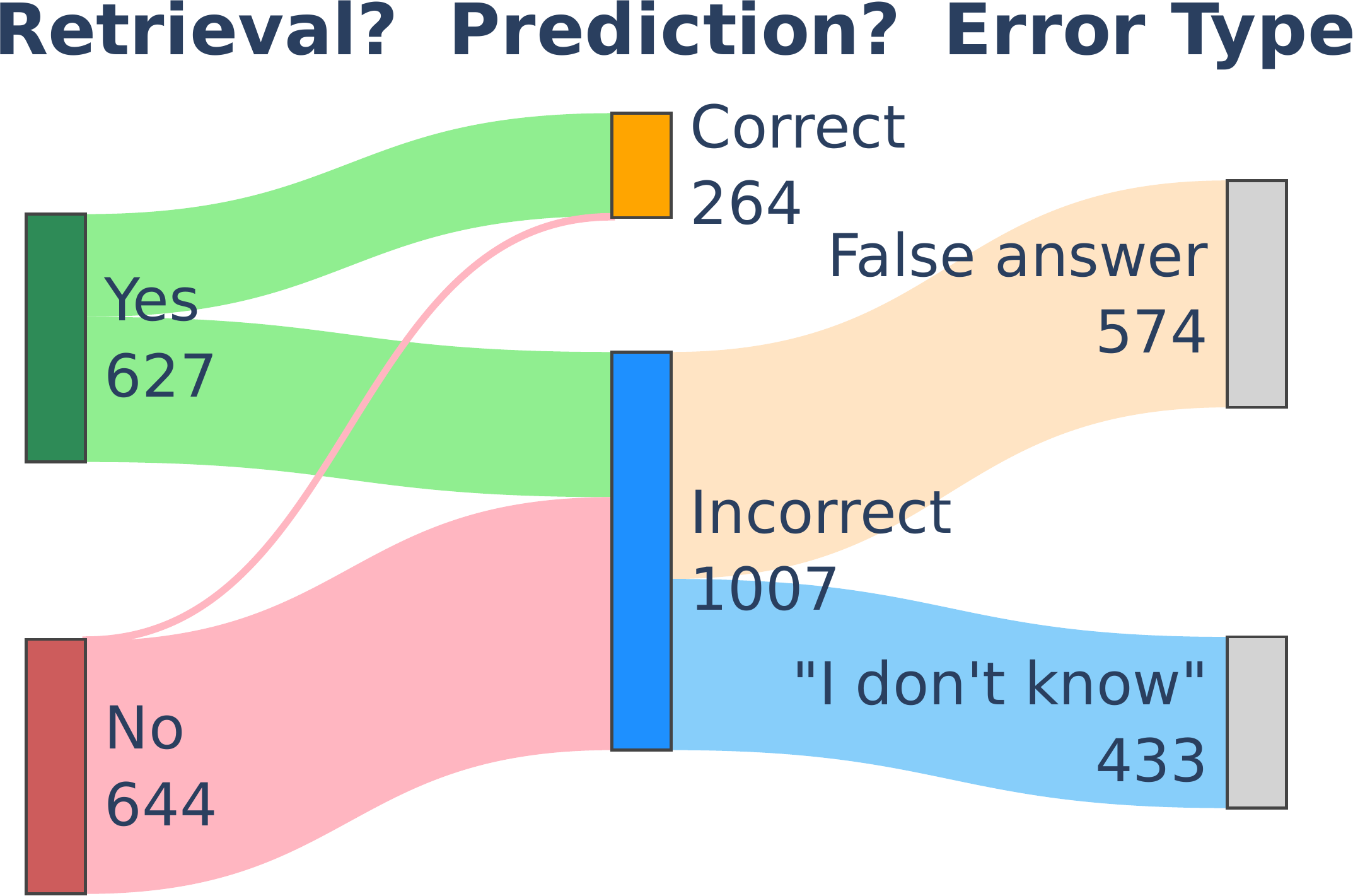}
    % \caption{
    % \footnotesize {
    % Error analysis for GPT-3.5 using Vanilla prompting
    % }}
\end{subfigure}
\end{tabular}
\caption{
% \footnotesize
\textbf{Above}: QA accuracy on our \texttt{RetrievalQA} w/, w/o retrieval, and adaptive retrieval. We set threshold $t=0.5$ for \textit{calibration-based} Self-RAG \citep{asai2023selfrag} and use \textit{model-based} \textbf{Vanilla} prompting for others (\cref{sec_preliminary}).
We find that
Self-RAG requires threshold tuning to balance QA performance and retrieval efficiency, while vanilla prompting is insufficient in guiding LLMs to make reliable retrieval decisions (\cref{sec_initial_results}).
\textbf{Below}:
an error analysis for GPT-3.5.
At least half of the time, GPT-3.5 is unaware that it needs retrieval (\ie, \colorbox{lightpink!100}{Red} area, \cref{sec_error_analysis}).
% We focus on reducing the \colorbox{lightpink!100}{Red} and \colorbox{lightskyblue!100}{Blue} areas.
}
\label{fig_sankey_gpt35}
\end{figure}
%%%%%%%%%%%% motivation figure %%%%%%%%%%%%

% To mitigate the aforementioned limitations of RAG, 
To alleviate the limitations of RAG mentioned above,
recent studies advocate for 
% \textit{adaptive retrieval},
\textbf{adaptive RAG (ARAG)},
which dynamically determines retrieval necessity and relies only on LLMs' parametric knowledge when deemed unnecessary \citep{feng2023trends}.
However, the effectiveness of these methods is understudied, as there is no suitable benchmark and evaluation.
ARAG approaches can be categorized into \textit{calibration-based} and \textit{model-based} judgement.
Calibration-based methods \citep{mallen-etal-2023-trust, jiang-etal-2023-active, asai2023selfrag}, while effective, trigger retrieval only when a metric surpasses a pre-defined threshold.
% While promising, these methods are primarily based on calibration, where retrieval triggers only when a metric surpasses a pre-defined threshold.
For example, \citet{mallen-etal-2023-trust} heuristically retrieve when the popularity of an entity on Wikipedia is below a certain threshold;
\citet{jiang-etal-2023-active} trigger retrieval if any token in the temporarily generated sentence has low confidence.
Clearly, these ad-hoc calibration-based methods are suboptimal, as we need to tune thresholds for different datasets and models to 
% maximize task performance while minimizing inference overheads.
balance task performance and inference overheads.
To obviate the hyperparameter threshold, model-based methods \citep{feng2023knowledge, ren2023investigating} directly prompt LLMs for retrieval decisions, given the observation that LLMs can acknowledge their knowledge boundaries to some extent \citep{kadavath2022language, yin-etal-2023-large}.
These methods undergo separate evaluations, and their effectiveness remains ambiguous due to the limited scope of the assessments.

% Given an input query, do LLMs know whether they need to retrieve external resources to answer correctly?
% In this paper, we investigate to what extent LLMs know whether they need to retrieve external resources to answer a given query correctly through the perspective of in-context learning (ICL; \citealt{NEURIPS2020_1457c0d6}).  
In this paper, we investigate to what extent LLMs can perform calibration-free adaptive retrieval 
% via in-context learning (ICL; \citealt{NEURIPS2020_1457c0d6}). 
via prompting.
To answer this question, we need to evaluate whether LLMs retrieve \textit{only} when necessary.
This requests a benchmark that distinguishes between questions that can be answered using LLMs' parametric knowledge and those that require external information through retrieval.
Nevertheless, commonly used open-domain QA datasets \citep{rajpurkar-etal-2016-squad, joshi-etal-2017-triviaqa, kwiatkowski-etal-2019-natural, mallen-etal-2023-trust} fail to fulfil this purpose, as various LLMs have distinct sizes and levels of pre-trained knowledge, making them inadequately assess the necessity of external retrieval for LLMs.
% \zihan{For example, LLmaa can understan}{}

% To answer this question, we first need a benchmark that distinguishes between questions that can be answered using LLMs' parametric knowledge and those that require external information through retrieval.
% Nevertheless, commonly used open-domain QA datasets \citep{rajpurkar-etal-2016-squad, joshi-etal-2017-triviaqa, kwiatkowski-etal-2019-natural, mallen-etal-2023-trust} inadequately assess the necessity of external retrieval for LLMs, as various models have distinct sizes and levels of pre-trained knowledge.
% It is inherently intractable to 
% For example, As shown in \zihan{TODO: Fig.1}{}.

% To fill this gap, we create \textbf{\texttt{RetrievalQA}}, a short-form QA dataset with 1,271 questions, covering new world and long-tail knowledge and spanning diverse topics.
To fill this gap, we create \textbf{\texttt{RetrievalQA}}, a short-form QA dataset, covering new world and long-tail knowledge and spanning diverse topics.
We ensure the knowledge necessary to answer the questions is absent from LLMs. Therefore, LLMs must truthfully decide whether to retrieve to be able to answer the questions correctly. 
\texttt{RetrievalQA} enables us to evaluate the effectiveness of ARAG approaches, an aspect predominantly overlooked in prior studies and recent RAG evaluation systems \citep{chen2023benchmarking, saadfalcon2023ares, es2023ragas}, which focus only on task performance, the relevance of retrieval context or the faithfulness of answers.

% Furthermore, the existing evaluation of RAG systems primarily focuses on downstream task performance \citep{asai2023selfrag} or retrieval context relevance and answer faithfulness \citep{chen2023benchmarking, saadfalcon2023ares, es2023ragas} while largely overlooking the necessity of retrieval.
% However, it is inherently intractable
% Some works have done ..., but they don't test...., we from another angle...
% An when to retrieval \citep{asai-etal-2023-retrieval}
% It is possible that different LLMs contains different knowledge, thus require different retrieval rate, however, we show that GPT3.5/4 are over-confident with their abilities.
% Furthermore, the existing evaluation of RAG systems primarily focuses on downstream task performance \citep{asai2023selfrag} or retrieval context relevance and answer faithfulness \citep{chen2023benchmarking, saadfalcon2023ares, es2023ragas} while largely overlooking the necessity of retrieval.
% Given a query, can LLMs judge the necessity of retrieval? Existing methods Decision may not always be optimal

Using \texttt{RetrievalQA} as a testbed, we benchmark both calibration-based and model-based methods with varying sizes of LLMs.
% (Fig.\ref{fig_sankey_gpt35}). 
As shown in Fig.\ref{fig_sankey_gpt35},
we find calibration-based Self-RAG requires threshold tuning to balance QA performance and retrieval efficiency, while vanilla prompting is insufficient in guiding LLMs to make reliable retrieval decisions.
As an initial effort, we propose \textbf{T}ime-\textbf{A}ware \textbf{A}daptive \textbf{RE}trieval (\textbf{TA-ARE}), a simple yet effective method to improve ARAG via in-context learning (ICL; \citealt{NEURIPS2020_1457c0d6}), obviating the need for calibration or additional training.

To sum up, this paper makes the following contributions: 
\textbf{\circled{1}} we create a new dataset \texttt{RetrievalQA} to assess ARAG for short-form open-domain QA;
\textbf{\circled{2}} 
we benchmark existing methods and conduct extensive analysis,
finding that vanilla prompting is insufficient in guiding LLMs to make reliable retrieval decisions;
\textbf{\circled{3}} we then propose TA-ARE, a simple yet effective method to help LLMs assess the necessity of retrieval without calibration or additional training.

\section{Dataset Construction}
\label{sec_dataset_construction}

% To facilitate our investigation, we construct \textbf{\texttt{RetrievalQA}}, a short-form open-domain QA dataset specifically crafted to assess the capacity of LLMs in discerning the need for external resource retrieval.

\paragraph{Data collection.}
Inspired by \citet{zhuang2023toolqa}, we aim to collect data such that the knowledge necessary to answer the questions is absent from LLMs. Therefore, LLMs must consult external resources to answer correctly.
Specifically, we mainly collect data from two categories:
\textbf{\circled{1} New world knowledge} that is out of the scope of the LLMs' pre-training corpora. LLMs are static after training and can quickly be outdated due to the ever-changing world \citep{zhang-etal-2023-large}. 
To ensure the knowledge is novel to most LLMs, we select 397 QA pairs ranging from 1 October 2023 to 12 January 2024 from RealTimeQA \citep{kasai2023realtime}. These data comprise weekly quizzes extracted from news websites, encompassing broad topics, including politics, business, and entertainment.  
In addition,
% Additionally,
we collect 127 fast-changing questions from FreshQA \citep{vu2023freshllms}, where the answers may change frequently, thereby challenging LLMs' parametric memorization.
\textbf{\circled{2} Long-tail knowledge} that is rarely learned during pre-training. Previous studies \citep{pmlr-v202-kandpal23a} have shown that LLMs struggle to learn less common knowledge and perform poorly without the help of retrieval.
% We use the long-tail subset of PopQA \citep{mallen-etal-2023-trust}, which consists of 1,399 rare entity queries whose monthly Wikipedia page views are less than 100.
Following \citet{asai2023selfrag},
we use the long-tail subset of PopQA \citep{mallen-etal-2023-trust}, which consists of 1,399 rare entity queries with monthly Wikipedia page views below 100,
and the test split of unfiltered TriviaQA \citep{joshi-etal-2017-triviaqa}, which has 7,313 factual QA pairs.
Lastly, we collect 100 personal agenda questions from ToolQA \citep{zhuang2023toolqa}, which are synthesized with virtual names and events.

\paragraph{Filtering out answerable questions.}
As discussed in \cref{sec_introduction}, to ensure the questions cannot be answered without external knowledge, we conduct strict filtering.
To save manual work, we prompt GPT-4 for answers
% \footnote{\texttt{gpt-4-turbo-preview}} 
in a closed-book QA setting without access to external knowledge (see prompt template Fig.\ref{fig_prompt_no_retrieval} in Appendix).
Then, we calculate the token-level F1 scores \citep{rajpurkar-etal-2016-squad} and remove questions that have shared tokens between the prediction and the ground truth, \ie, only keep questions with $\text{F1} = 0$.
Our rationale is that if state-of-the-art GPT-4 cannot answer correctly without retrieval, weaker LLMs are also highly likely to fail.
Finally, after filtering, we have obtained 1,271 out of 9,336 questions, covering new world and long-tail knowledge and spanning diverse topics.
To avoid potential bias in the evaluation towards methods that retrieve more often, we additionally collect 1,514 questions that can be answered using GPT-2’s parametric knowledge from the discard set. More details are in the Appendix \ref{append_dataset_construction}.

We conduct a sanity check using various sizes of LLMs in Fig.\ref{fig_sankey_gpt35} and in Appendix \ref{append_no_retrieval}, showing that \texttt{RetrievalQA} is extremely hard for all models without access to external knowledge.
We present detailed data statistics in Table \ref{tab_data_statistics} and examples of the data in Table \ref{tab_data_instances}.
% The detailed data statistics are in Table \ref{tab_data_statistics}.

% \section{Method}
% \label{sec_method}

% \section{Preliminary Investigation}
\section{\texttt{RetrievalQA} Challenges Adaptive RAG}
\label{sec_preliminary}

% In this section, we formalize adaptive retrieval-augmented generation for open-domain QA tasks.
In this section, we formalize ARAG for open-domain QA tasks and evaluate existing adaptive approaches on \texttt{RetrievalQA}.

\subsection{Standard \& Adaptive RAG for QA}
\label{sec_adaptive_rag_method}
% \paragraph{Standard RAG for QA.}

\paragraph{Standard RAG.}
Given a question $x$, a retriever $\mathcal{R}$ , and an external document corpus 
 % $\mathcal{D}=\left\{d_i\right\}_{i=1}^m$ 
$\mathcal{D}$ 
such as Wikipedia, the retriever first retrieves a list of relevant documents $\mathcal{D}_x = \mathcal{R}(x)$,
then an LLM needs to generate answer $y = \text{LLM}(I, \mathcal{D}_x, x)$ conditioned on a prompt instruction $I$, retrieved documents $\mathcal{D}_x$, and the question $x$.
% , where $[\cdot, \cdot]$ means concatenation.

% \subsection{Adaptive RAG for QA}

\paragraph{Adaptive RAG.}
Standard RAG always retrieves regardless of the input question, while adaptive retrieval only retrieves when necessary.
\textbf{Calibration-based} methods generally introduce a pre-defined hyperparameter $t$ and only do retrieval when a metric surpasses $t$:
% {\small
\begin{align*}
% \normalsize
% \small
y =\left\{\begin{array}{l}
\text{LLM}(I, \mathcal{D}_{x}, x),  \text { metric } \geq t \\
\text{LLM}(I, x), \quad \ \ \ \text { otherwise }
\end{array}\right.
\end{align*}
% }
\textbf{Baselines.} While our primary focus is model-based methods, we also evaluate the most recent state-of-the-art Self-RAG (7B) \citep{asai2023selfrag}.
Self-RAG fine-tunes Llama-2 using special reflection tokens to allow the model to introspect its outputs. 
The model activates retrieval when the probabilities of the generated special tokens exceed a threshold.
% \paragraph{Model-based.} Following \citet{feng2023knowledge, ren2023investigating}, we instruct LLMs to let themselves decide whether to retrieve, obviating the threshold.

For \textbf{Model-based} methods,
we follow \citet{feng2023knowledge} and \citet{ren2023investigating} to instruct LLMs to decide whether to retrieve via prompting, obviating the threshold.
Specifically, we ask a yes/no question: $r = \text{LLM}(I_{\text{vanilla}}, x)$, where $I_{\text{vanilla}}=$ 
{\small
\texttt{"Given a question, determine whether you need to retrieve ... answer [Yes] or [No]"}.
}
Retrieval is performed only when LLMs answer yes. 
We denote this as \textbf{Vanilla} prompting:
% In \cref{sec_icare}, we propose a simple yet effective approach to improve ARAG.
\begin{align*}
% \small
y =\left\{\begin{array}{l}
\text{LLM}(I, \mathcal{D}_{x}, x), \ r = \text{Yes} \\
\text{LLM}(I, x), \quad \ \ \ \text { otherwise }
\end{array}\right.
\end{align*}
\textbf{Baselines.} We evaluate strong instruction-tuned models with a varying scale of model size: TinyLlama (1.1B; \citealt{zhang2024tinyllama}), 
Phi-2
% \footnote{We acknowledge that Phi-2 has not been instruction fine-tuned; however, we find it performs decently well in understanding instructions.}
(2.7B;
 % \citep{gunasekar2023textbooks, li2023textbooks}, 
  \citealt{li2023textbooks}), 
Llama-2 (7B; \citealt{touvron2023llama}), GPT-3.5 \citep{OpenAI_chatgpt2022}, and GPT-4 \citep{openai2023gpt4}.

%%%%%%%%%%%%%%%%%%%%%% vanilla/TA-ARE %%%%%%%%%%%%%%%%%%%%%%
\begin{table}[t]
% \small
% \footnotesize
\scriptsize
\centering
\begin{tabular}{lCCC}
\toprule
\multirow{2}[3]{*}{\B Baselines (1,271)}   & \multicolumn{2}{c}{\textbf{Adaptive Retrieval}} & \textbf{\makecell{Always \\ Retrieval}} \\ 

\cmidrule(lr){2-4} & \text{Retrieval} & \text{Match}  & \text{Match}                  \\
\midrule
% \noalign{\vskip 1.5mm}
% {\textbf{\textit{Calibration-based}}} & & & \\

%  & \multicolumn{3}{c}{\textbf{\textit{Calibration-based}}} \\

% Self-RAG (7B) $_{\texttt{t=0.25}}$ & 100.0 & 31.9 & \multirow{4}{*}{$31.9$} \\
% Self-RAG (7B) $_{\texttt{t=0.5}}$  & 23.0  & 10.6  &   \\
% Self-RAG (7B) $_{\texttt{t=0.75}}$ & 0.0  & 6.0   \\
% Self-RAG (7B) $_{\texttt{t=None}}$ & 0.4  & 6.0  \\
% \midrule

& \multicolumn{3}{c}{\textbf{\textit{Calibration-based}}} \\

Self-RAG (7B) & & & \\
\quad $t=0.25$ & 100.0 & 31.9 & \multirow{4}{*}{$31.9$} \\
\quad $t=0.5$   & 23.0  & 10.6  &   \\
\quad $t=0.75$  & 0.0  & 6.0   \\
\quad $t=\texttt{None}$ & 0.4  & 6.0  \\

\midrule

% \cdashline{1-4}

 & \multicolumn{3}{c}{\textbf{\textit{Model-based}}} \\
\cellcolor{lightgray!25} \textbf{\textit{Vanilla}} \ \cref{sec_preliminary}  & \cellcolor{lightgray!25} & \cellcolor{lightgray!25} & \cellcolor{lightgray!25} \\

TinyLlama (1.1B) & 39.1 & 14.7  & 28.2   \\
Phi-2 (2.7B)     & \mathbf{94.1} & \underline{35.0}  & 36.4   \\
Llama-2 (7B)     & \underline{80.3} & 26.1  & 36.0   \\
% Self-RAG (7B) $_{\texttt{t=None}}$ & 0.4  & 6.0 & 31.9 \\
GPT-3.5          & 49.3 & 20.8  & \underline{38.2}    \\
GPT-4*          & 67.6 & \mathbf{37.6}  & \mathbf{46.0}  \\

% \midrule
% \cdashedline{1-4}
% \cmidrule(l){1-4}
% \midrule[dashed]
\hdashline

%  & \multicolumn{3}{c}{\textbf{\textit{Model-based}}} \\
\cellcolor{lightgray!25} \textbf{\textit{Ours TA-ARE}} \ \cref{sec_icare}  & \cellcolor{lightgray!25} & \cellcolor{lightgray!25} & \cellcolor{lightgray!25} \\

TinyLlama (1.1B) & 54.1 \text{\scriptsize{(\G{+15.0})}} & 19.0 \text{\scriptsize{(\G{+4.3})}} & 28.2 \\
Phi-2 (2.7B)     & \mathbf{95.5} \text{\scriptsize{(\G{+1.4})}} & \underline{36.0} \text{\scriptsize{(\G{+1.0})}} & 36.4 \\
Llama-2 (7B)     & 86.0 \text{\scriptsize{(\G{+5.7})}} & 30.7 \text{\scriptsize{(\G{+4.6})}}  & 36.0 \\
% Self-RAG (7B) $_{\texttt{t=None}}$ & 0.4  & 6.0 & 31.9 \\
GPT-3.5          & \underline{86.3} \text{\scriptsize{(\G{+37.0})}} & 35.8 \text{\scriptsize{(\G{+15.0})}}  &  \underline{38.2}  \\
GPT-4*            & 83.2  \text{\scriptsize{(\G{+15.6})}} & \mathbf{46.4}  \text{\scriptsize{(\G{+8.8})}} &  \mathbf{46.0} \\  

% \midrule
Average gain & \G{\text{+}14.9} & \G{\text{+}6.7} & \text{--} \\

\bottomrule
\end{tabular}
\caption{
% \footnotesize
% \textbf{Vanilla} prompting 
Retrieval and match accuracy on \texttt{RetrievalQA}. 
% For Self-RAG, we set the retrieval threshold $t = [0.25, 0.5, 0.75, \texttt{None}]$. \texttt{None} means the model itself decides when to retrieve.
* indicates using 250 examples for testing to reduce API costs.
Best scores in \textbf{Bold} and second best in \underline{underline}.
}
\label{tab_vanilla_results}
\end{table}
%%%%%%%%%%%%%%%%%%%%%%  vanilla/TA-ARE %%%%%%%%%%%%%%%%%%%%%%

\subsection{Experiment Setup}
% \section{Experiment}

% Using \texttt{RetrievalQA} as a testbed, we qualify adaptive retrieval approaches.

\paragraph{Evaluation Metric.}
We use \textit{retrieval} accuracy to evaluate how well LLMs can perform adaptive retrieval. Since all questions in our dataset need retrieval, the higher the retrieval accuracy, the more effective the method. 
Following \citet{schick2023toolformer, mallen-etal-2023-trust, asai2023selfrag}, we evaluate QA performance using \textit{match} accuracy, which measures whether gold answers are included in the model predictions instead of strict exact matching. 
% We additionally report token-level F1 scores in the Appendix.
% We evaluate whether LLMs can make correct predictions of whether retrieval is required using accuracy.
% We also calculate the token reduction rate.

\paragraph{Implementation details.}
% For fair comparisons, we use the same setting following Self-RAG for all experiments. The detailed hyperparameters are summarized in Table \ref{tab_hyperparameters}.
% For Self-RAG, we set the retrieval threshold $t = [0.25, 0.5, 0.75, \texttt{None}]$. Lower thresholds encourage more frequent retrieval, while \texttt{None} means the model itself decides when to retrieve by generating the specific \texttt{[Retrieval]} token.
% Since the quality of the retrieved documents is not the focus of this paper, we use the off-the-shelf Contriever \citep{izacard2022unsupervised} and author-provided top-5 documents extracted from Wikipedia where possible for long-tail knowledge questions. 
% For questions from ToolQA, we use the author-provided vector database for retrieval of synthesized agendas.
% Otherwise, we use top-5 documents returned by Google search\footnote{We use \href{https://serpapi.com/}{SerpApi} for Google search.} for new world knowledge questions.
% To reduce API costs, for GPT-4, we randomly select 50 data instances from each source for evaluation, resulting in 250 questions.
% We ask LLMs to respond \texttt{"I don't know"} if they cannot answer the question.

For Self-RAG, we set the retrieval threshold $t = [0.25, 0.5, 0.75, \texttt{None}]$. Lower thresholds encourage more frequent retrieval, while \texttt{None} means the model itself decides when to retrieve by generating the specific \texttt{[Retrieval]} token.
% We use top-5 retrieved documents, either using Contriever \citep{izacard2022unsupervised} or Google search.
Since the quality of the retrieved documents is not the focus of this paper, for long-tail knowledge questions, we use the off-the-shelf Contriever \citep{izacard2022unsupervised} and author-provided top-5 documents extracted from Wikipedia where possible. 
For questions from ToolQA, we use the author-provided vector database for retrieval of synthesized agendas.
Otherwise, we use top-5 documents returned by Google search
% \footnote{We use \href{https://serpapi.com/}{SerpApi} for Google search.} 
for new world knowledge questions.
To reduce API costs, for GPT-4, we randomly select 50 data instances from each source for evaluation, resulting in 250 questions.
We ask LLMs to respond \texttt{"I don't know"} if they cannot answer the question.
Due to page limitation, we primarily evaluate the 1,271 questions that need retrieval in the main body and provide the overall results in the Appendix \ref{appendix_overall_results}.
More implementation details and prompt templates are in Appendix \ref{append_implementation_details} and \ref{appendix_prompt}.

\subsection{Results}
\label{sec_initial_results}

% \paragraph{Main results.}
Table \ref{tab_vanilla_results} (top \& middle) shows the retrieval accuracy and answer match accuracy for calibration-based and model-based methods. We also present the results of standard RAG, \textit{Always Retrieval}, which can be seen as the upper bound of the baselines. We observe that:
\textbf{\circled{1} RAG generally improves QA performance.} As the knowledge necessary to answer the questions is not present in LLMs, the more frequently retrieval occurs, the higher the answer accuracy becomes for all models.
However, GPT-4 possesses the highest QA accuracy despite retrieving only $67.6\%$ of the time, indicating that fully utilizing retrieved context is also crucial for generating correct answers \citep{asai-etal-2023-retrieval}.
\textbf{\circled{2} The effectiveness of Self-RAG largely depends on threshold tuning.} As shown in Table \ref{tab_vanilla_results} (top), Self-RAG achieves high performance when setting a low retrieval threshold ($t=0.25$) while never retrieving when the threshold is high ($t=0.75$). 
This indicates that calibration-based methods require threshold tuning to find the best trade-off between task performance and retrieval efficiency. 
\textbf{\circled{3} The effectiveness of vanilla prompting varies and does not scale with model sizes.} 
Surprisingly, Table \ref{tab_vanilla_results} (middle) shows larger models (GPT-3.5/4) perform worse than smaller yet strong models (Phi-2/Llama-2) in retrieval accuracy, suggesting that LLMs possess a certain degree of ability to perceive their knowledge boundaries \citep{yin-etal-2023-large, ren2023investigating}.
% However, Vanilla prompting could not fully elicit this ability.
Yet, vanilla prompting is insufficient in guiding LLMs to make reliable retrieval decisions.

\subsection{Error Analysis}
\label{sec_error_analysis}

\begin{figure}[t]
\centering
\includegraphics[width=\columnwidth]{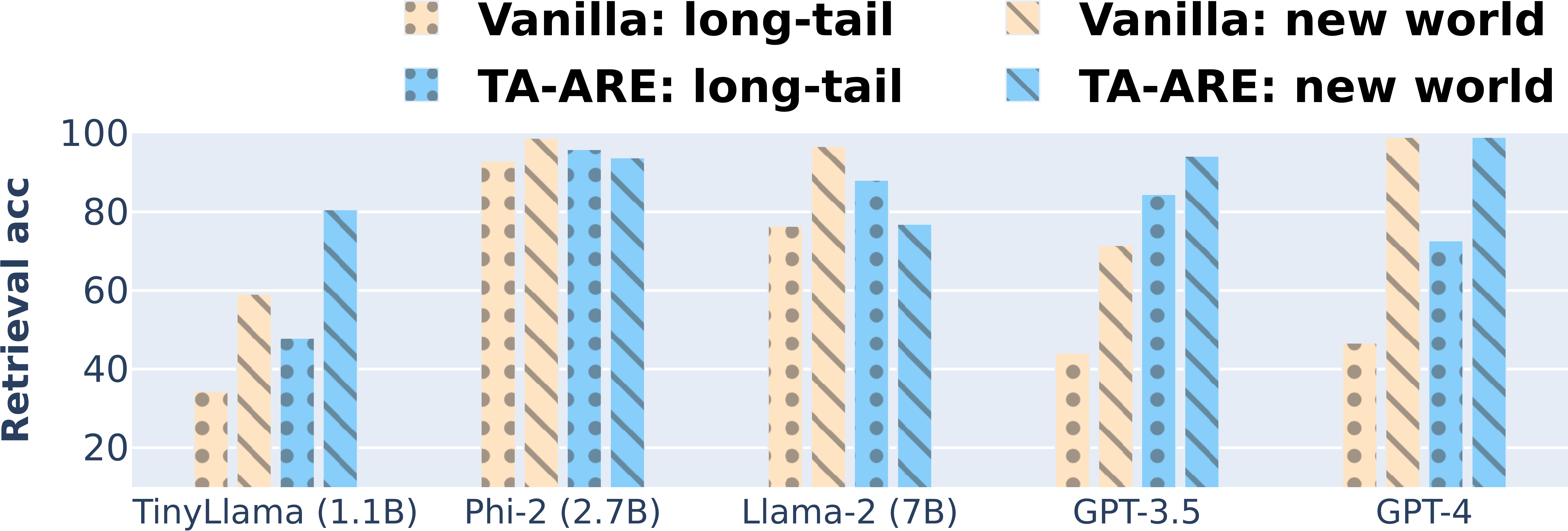}
\caption{
% \footnotesize
Retrieval accuracy between \textit{long-tail} vs. \textit{new world} knowledge (\ie, dotted vs. slash) using \textbf{Vanilla} and ours \textbf{TA-ARE} (\ie, yellow vs. blue).
}
\label{fig_compare_longtail_new}
\end{figure}

% \paragraph{Error analysis.}
To investigate why vanilla prompting performs poorly for ARAG, we conduct an error analysis for GPT-3.5 and plot Fig.\ref{fig_sankey_gpt35}. 
% More results in Appendix \ref{append_error_analysis}.
% \zihan{More results in Appendix}{}
% \colorbox{bisque!100}{yellow} also contains the cases when LLMs conduct retrieval but cannot fully utilize the context, or the context is noisy and irrelevant, thus making the wrong predictions. However, this is not the focus of this paper.
\colorbox{lightpink!100}{Red} area indicates more than half of the time, GPT-3.5 overconfidently perceives no external information is required to answer the questions, leading to mostly incorrect predictions. 
Conversely, \colorbox{lightskyblue!100}{Blue} area shows that, without additional information, GPT-3.5 \textit{"knows"} it does not know the answer, therefore responding \texttt{"I don't know"}.
Together, 
% this reveals that LLMs possess the ability of self-knowledge, 
this reveals that LLMs can potentially discern the need for resource retrieval.
% but how can we elicit this ability and reduce the \colorbox{lightpink!100}{Red} and \colorbox{lightskyblue!100}{blue} areas?
% We further compare the retrieval accuracy between long-tail and new world knowledge in Fig.\ref{fig_compare_longtail_new} (yellow), finding that all LLMs have
% We further find that all LLMs can better acknowledge their lack of new world knowledge, while weak in long-tail questions, as in Fig.\ref{fig_compare_longtail_new} (yellow).
We further find in Fig.\ref{fig_compare_longtail_new} that all LLMs can better recognize their lack of knowledge about the new world, leading them to actively request retrieval. However, they tend to be weak in handling long-tail questions, as depicted in Fig.\ref{fig_compare_longtail_new} (yellow).

%%%% Error analysis: long tail vs. new world knowledge %%%%

% %%%% Error analysis: long tail vs. new world knowledge %%%%
% \begin{figure}[ht]
% \centering
% \includegraphics[width=0.4\columnwidth]{figs/ablation_example_k-cropped.pdf}
% \caption{
% Effect of different numbers of demonstrations. Averaged for all models. 
% % At least 433/644 ()
% }
% \label{fig_ablation_k}
% \end{figure}
% %%%% Error analysis: long tail vs. new world knowledge %%%%

% %%%%% prompt template ablation %%%%%%%%%%
% \begin{table}[ht]
% % \small
% \footnotesize
% % \scriptsize
% \centering
% \begin{tabular}{lccCC}
% \toprule
%  & \text{\B Time} & \text{\B Example} & \text{\B Avg. Retrieval} & \text{\B Avg. Match} \\
% \midrule
% % vanilla
% 1 & &  & 65.8 & 24.8 \\
% % no_fewshot_time
% 2 & \ding{52} &  & 72.4 & 27.0  \\
% % fewshot_no_time
% 3 & & \ding{52} & 78.9 & 29.3  \\
% % fewshot_time
% 4 & \ding{52} & \ding{52}  & 80.6 & 31.1 \\

% \bottomrule
% \end{tabular}
% \caption{
% % \footnotesize 
% Ablation study for current date and demonstration examples. Results are averaged for all models.
% }
% \label{tab_prompt_ablation_results}
% \end{table}
% %%%%% prompt template ablation %%%%%%%%%%

% \paragraph{Retrieval}
% For new world knowledge, we 

\section{Improving Adaptive RAG Prompting}
\label{sec_icare}

This section presents an improved model-based ARAG method and evaluates its effectiveness.

\subsection{Method}

% \paragraph{Method.}
Based on our findings in \cref{sec_error_analysis}, we propose 
 \textbf{T}ime-\textbf{A}ware \textbf{A}daptive \textbf{RE}trieval via ICL (\textbf{TA-ARE}),
a simple yet effective method to improve ARAG without calibration or additional training.
 % obviating the need for calibration or additional training.
Given that new world knowledge questions often contain time-sensitive information (\eg, "last week", "recent"), 
% \citep{shang-etal-2022-improving},
we include 
{\small
"\texttt{Today is current\_date()}"
}
in the instruction to enhance models' awareness of time.
For long-tail knowledge, we use SimCSE \citep{gao-etal-2021-simcse} to select top-2 semantically closest long-tail questions answered incorrectly from the discarded set in \cref{sec_dataset_construction}, denoted as \texttt{[Yes]} demonstrations.
For \texttt{[No]} demonstrations, we manually create another two questions, ensuring no extra information is required for most LLMs to answer 
(\eg, 
{\small
\texttt{What is the capital of France?}
}).

%%%% Error analysis: long tail vs. new world knowledge %%%%
\begin{figure}[t]
\centering
\includegraphics[width=0.7\columnwidth]{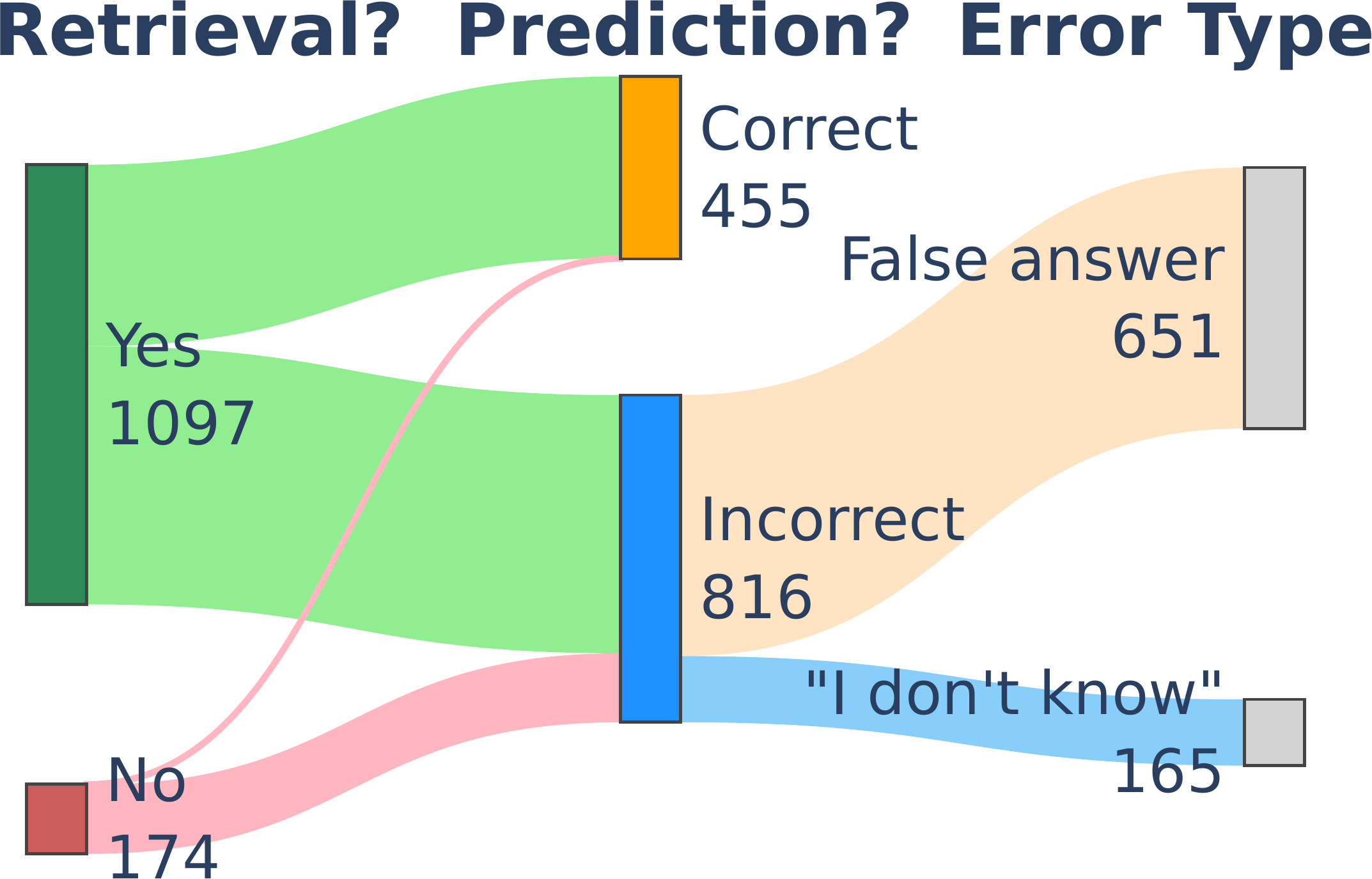}
\caption{
Error analysis of ours \textbf{TA-ART} for GPT-3.5. Compared to Fig.\ref{fig_sankey_gpt35}, we can see that the areas of \colorbox{lightpink!100}{Red} and \colorbox{lightskyblue!100}{Blue} significantly reduce, indicating that GPT-3.5 has improved awareness of when it needs retrieval.
}
\label{fig_sankey_gpt35_ours}
\end{figure}
%%%% Error analysis: long tail vs. new world knowledge %%%%

%%%%% prompt template ablation %%%%%%%%%%
\begin{table}[ht]
% \small
\footnotesize
% \scriptsize
\centering
\begin{tabular}{lccCC}
\toprule
 & \text{\B Time} & \text{\B Example} & \text{\B Avg. Retrieval} & \text{\B Avg. Match} \\
\midrule
% vanilla
1 & &  & 65.8 & 24.8 \\
% no_fewshot_time
2 & \ding{52} &  & 72.4 & 27.0  \\
% fewshot_no_time
3 & & \ding{52} & 78.9 & 29.3  \\
% fewshot_time
4 & \ding{52} & \ding{52}  & 80.6 & 31.1 \\

\bottomrule
\end{tabular}
\caption{
% \footnotesize 
Ablation study for current date and demonstration examples. Results are averaged for all models.
}
\label{tab_prompt_ablation_results}
\end{table}
%%%%% prompt template ablation %%%%%%%%%%

%%%% Error analysis: long tail vs. new world knowledge %%%%
\begin{figure}[ht]
\centering
\includegraphics[width=0.5\columnwidth]{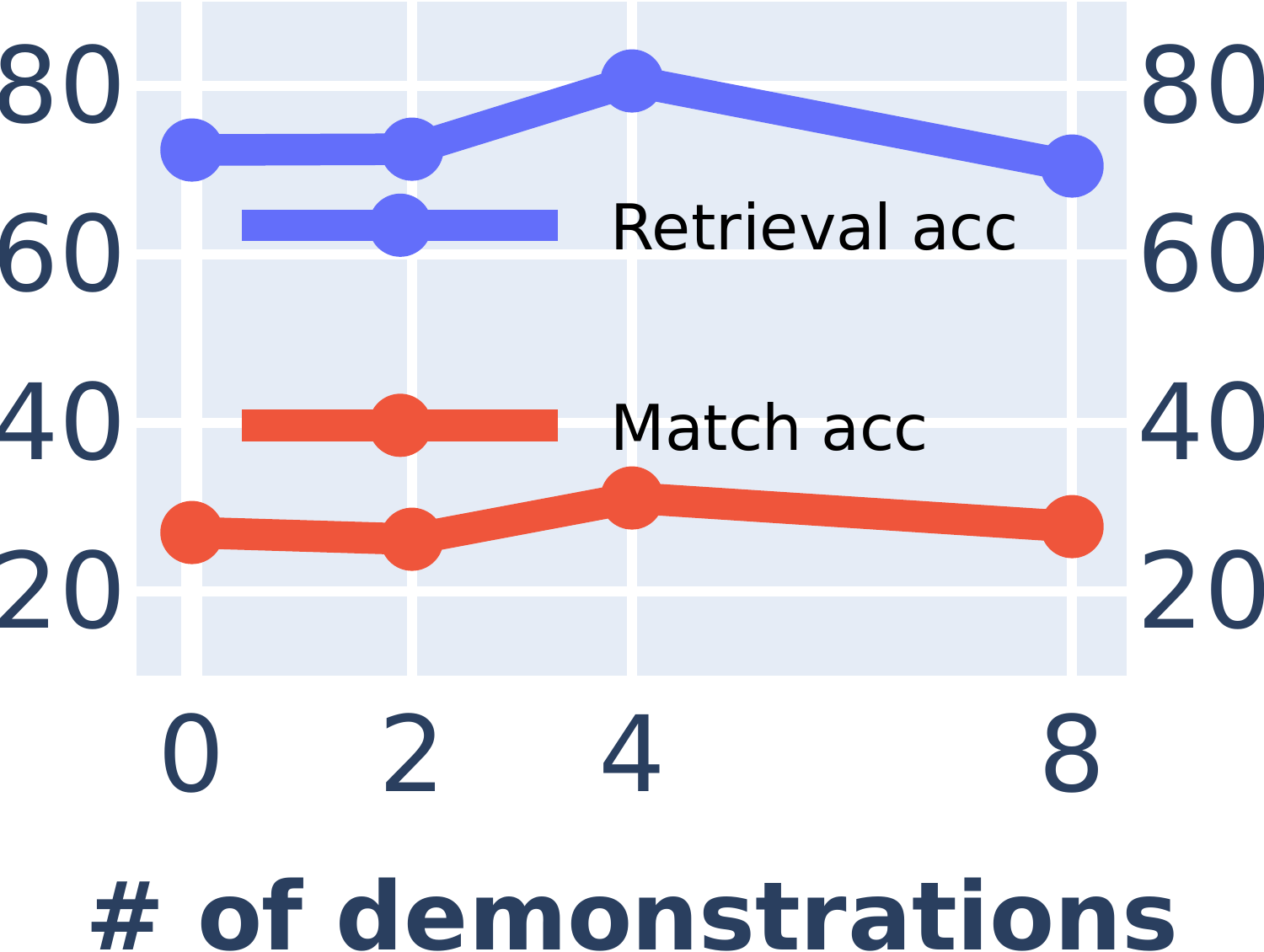}
\caption{
Effect of different numbers of demonstrations. Averaged for all models. 
% At least 433/644 ()
}
\label{fig_ablation_k}
\end{figure}
%%%% Error analysis: long tail vs. new world knowledge %%%%

\subsection{Results}

% \paragraph{Results \& Ablation.}
% As shown in Table \ref{tab_vanilla_results} and Fig.\ref{fig_compare_longtail_new}, 
% Table \ref{tab_vanilla_results} and Fig.\ref{fig_compare_longtail_new} shows 
% TA-ARE significantly improves all baselines, 
% with an average gain of $14.9\%$ and $6.7\%$ for retrieval and QA accuracy, respectively.
Table \ref{tab_vanilla_results} shows 
TA-ARE significantly improves all baselines, 
with an average gain of $14.9\%$ and $6.7\%$ for retrieval and QA accuracy, respectively.
Fig.\ref{fig_compare_longtail_new} illustrates the improvement for all long-tail questions and most new world questions.
As shown in Fig.\ref{fig_sankey_gpt35_ours}, we plot the error analysis on GPT-3.5 using our proposed TA-ARE. Compared to Fig.\ref{fig_sankey_gpt35} which uses vanilla prompting, we can see that the areas of \colorbox{lightpink!100}{Red} and \colorbox{lightskyblue!100}{Blue} significantly reduce, 
% indicating that GPT-3.5 possesses the ability to discern whether to retrieve. 
indicating that GPT-3.5 has improved awareness of when it needs retrieval, 
demonstrating our approach successfully elicits this ability.

In addition, our plotting enables us to conduct fine-grained error analysis for RAG. 
We can see that part of the 
\colorbox{bisque!100}{LightYellow} area (when \texttt{Retrieval=Yes} and \texttt{Prediction=Incorrect}) generally represents two cases: First, the retrieved documents are noisy and might not contain relevant information to answer the questions. Thus, LLMs cannot make correct predictions; Second, the retrieved documents contain necessary information, but LLMs cannot fully utilize them and make correct predictions.
While this is out of the scope of this paper,  future works are required to make RAG systems more robust and effective \citep{asai-etal-2023-retrieval, yoran2023making}.

\subsection{Ablation Study}

The ablation studies in Table \ref{tab_prompt_ablation_results} validate the effectiveness of TA-ARE:
time awareness and relevant in-context demonstrations help LLMs decide the necessity of retrieval for new world and long-tail questions.
Table \ref{tab_fewshot_no_time_results} and Table \ref{tab_no_fewshot_time_results} in the Appendix further show the fine-grained performance of each model.
We further evaluate the number of in-context demonstrations in Fig.\ref{fig_ablation_k}, showing that 4 demonstrations, comprising 2 \texttt{[Yes]} and 2 \texttt{[No]} examples, have the best performance.

%%%%%%%%%%%%%%%%%%%%%%%%%%%%%%%%
%%% Since our data only contain questions that need retrieval, using 4 [Yes] examples will create bias, will encourage more retrieval
%%%%%%%%%%%%%%%%%%%%%%%%%%%%%

% We conduct ablation studies for TA-ARE in Table \ref{tab_prompt_ablation_results} and present detailed results in Appendix \zihan{TODO}{}.
% Row 1 indicates Vanilla prompting, while Row 4 represents our TA-ARE. Adding current date information helps models sense the time, which benefits discerning questions that contain new world information.
% Using in-context demonstrations further guides LLMs to understand the, especially for long-tail knowledge. 

\section{Conclusion}

This paper presents a new dataset \texttt{RetrievalQA} to assess adaptive RAG for short-form open-domain QA.
We find vanilla prompting is insufficient in guiding LLMs to make reliable retrieval decisions.
As an initial attempt,
% As a simple yet effective improvement, 
we propose TA-ARE, a simple yet effective method to help LLMs assess the necessity of retrieval, obviating the need for calibration or additional training.

% This paper study how well LLMs can perform calibration-free adaptive RAG via prompting. We construct a new dataset \texttt{RetrievalQA}

\section*{Limitations}

We identify the limitations of our work as follows:

\begin{itemize}
  \item We mainly collect data from existing data sources and use GPT-4 for filtering out answerable questions. While we have done preliminary human checking in Appendix \ref{append_no_retrieval}, it is possible that some questions in the dataset do not require additional information for LLMs to answer. Future work could develop advanced algorithms to do more efficient and rigour filtering.
  
  \item We primarily focus on short-form QA in this paper and do not assess long-form generation tasks. It should be noted that methods, including \citet{jiang-etal-2023-active, asai2023selfrag}, are capable of long-form generation tasks. Self-RAG can also perform sophisticated self-reflection, which goes beyond adaptive retrieval.
  
  \item We acknowledge that some of the retrieved documents may not contain the answers or the information needed to answer the questions. While improving retrieval relevance and accuracy is out of the scope of this paper, noisy context may interfere with LLMs and hurt the QA performance.

  \item While we find our prompt templates work well, we do not perform prompt tuning in this paper. We acknowledge that prompt templates can be sensitive to LLMs, and there are methods to find optimal prompts \citep{shin-etal-2020-autoprompt, deng-etal-2022-rlprompt}. We believe optimal prompts can be found and further improve performance. We leave this as a future work.

\end{itemize}

\section*{Ethical Statement}

The \texttt{RetrievalQA} dataset, meticulously curated to evaluate LLMs' self-awareness ability to decide when to retrieve external resources, is constructed only using publicly available data sources. 
We rigorously vetted the licenses of the five publicly available datasets for compliance, ensuring that all our research methodologies aligned with institutional, national, and global ethical standards.
We carefully examine the data to ensure no privacy concerns or violations. We do not collect any personally identifiable information. All data used in this paper is obtained following legal and ethical standards.
In addition, we adhere to the terms of use and policies of OpenAI and Meta.

\section*{Acknowledgements}

% This work is supported by TPG Telecom. 
This work is supported by the Australian Research Council DP240102349.
We thank anonymous reviewers for their valuable comments.

% Entries for the entire Anthology, followed by custom entries
\bibliography{anthology,custom}

\appendix

\section{Appendix}

\subsection{Data Sources}
\label{appendix_dataset}

% In this work, we collect data that are guaranteed cannot be answered without external information. 
% The dataset statistics are shown in Table \ref{tab_data_statistics}.
% The examples of data instances are in Table \ref{tab_data_instances}.
% We collect data from the following sources:
In this work, we collect data from the following sources:

\paragraph{RealTimeQA \citep{kasai2023realtime}}  A dynamic question-answering (QA) based on weekly-published news articles, which challenges static LLMs. We select data from 1 October 2023 to 12 January 2024. These data comprise weekly quizzes extracted from news websites, encompassing broad topics, including politics, business, and entertainment.  

 \paragraph{FreshQA \citep{vu2023freshllms}}   A QA benchmark with 600 questions that cover a wide range of questions and answer types. We use the fast-changing subset so that the knowledge memorized in LLMs can potentially be outdated, thus requiring external new information.

\paragraph{ToolQA \citep{zhuang2023toolqa}}  A benchmark to faithfully evaluate LLMs' ability to use external tools. We use questions from the Personal Agenda domain, which consists of 100 synthesized questions with virtual names and events.

\paragraph{PopQA \citep{mallen-etal-2023-trust}}  
An entity-centric open-domain QA dataset about entities with a wide variety of popularity. We use the long-tail subset of the data.

\paragraph{TriviaQA \citep{joshi-etal-2017-triviaqa}} 
A reading comprehension dataset containing question-answer-evidence triples. We follow \citet{asai2023selfrag} and use the test split of the unfiltered version.

\subsection{Details of \texttt{RetrievalQA} Construction}
\label{append_dataset_construction}

In this section, we discuss the details of our dataset construction. 
Our goal is to evaluate adaptive RAG (ARAG) methods and see how good they are at deciding when to retrieve. Therefore, we need the ground truth labels for each question's retrieval necessity. Ultimately, there are three kinds of questions here:

\begin{itemize}
    \item Case 1: for all LLMs, questions that can be answered using only their parametric knowledge
    \item Case 2: for all LLMs, questions that can \textit{not} be answered using only the parametric knowledge, therefore requiring external retrieval
    \item Case 3: questions that can be answered with their parametric knowledge for some models but can not be answered for some other models
\end{itemize} 

We do not consider Case 3 because those questions cannot fairly measure whether retrieval is required for different LLMs. For Case 1, it is not trivia to collect questions that can be answered only using the parametric knowledge of LLMs. This is because different LLMs have different levels of pre-trained knowledge, and it is hard to measure \citep{petroni-etal-2019-language, kadavath2022language}.
For example, given a question, GPT-3.5 may fail to answer and need the help of external knowledge, while Llama-2 may answer correctly using its own knowledge because it has seen the question in the training data. The pre-training corpora are sometimes unavailable, especially for proprietary models, and we can not guarantee that the collected questions can be 100\% answered with their own knowledge, as shown in the table below.

However, different from Cases 1 and 3, for Case 2, theoretically, it is possible to collect data that guarantees the knowledge to answer the questions is not present in the models. For instance, new world knowledge occurred after model training and long-tail knowledge that did not (or rarely) appear in the training corpora. Therefore, we collect 1,271 questions (Case 2) that are guaranteed cannot be answered without external information. The data collection process is detailed in \cref{sec_dataset_construction}.
The dataset statistics are shown in Table \ref{tab_data_statistics}.
The examples of data instances are in Table \ref{tab_data_instances}.

To avoid potential bias in the evaluation towards methods that retrieve more often, we additionally collect 1,514 questions (Case 1) that can be answered using GPT-2’s parametric knowledge from the discard set. Specifically, we use GPT-2 (small, 124M) in the zero-shot closed-book QA setting to evaluate the discard set. We only keep questions that can be answered using GPT-2’s parametric knowledge (when the loose match score = 1), assuming that larger and stronger LLMs are also highly likely to succeed if small and weak GPT-2 can answer correctly without retrieval. We also use the entire PopQA dataset (the rest of the long-tail split), which has 12,883 data instances that are more common on the web. We found that GPT-2 cannot answer any new-world questions from the discard set, which is reasonable.

%%%%%%%%%%%%%%%%%%%%%%%%%%%%%%%%%%%%%%%%%%%%%%%%%%
% Dataset V4 - total 1271
%%%%%%%%%%%%%%%%%%%%%%%%%%%%%%%%%%%%%%%%%%%%%%%%%%
\begingroup
\begin{table*}[ht]
\small
\centering
\begin{tabular}{ccCCCCC}
\toprule
\textbf{Category} & \textbf{Data Source} & \text{\textbf{\makecell{\# Original}}} & \text{\textbf{\makecell{\# After \\ Filtering}}} & \text{\textbf{\makecell{\# Avg. \\ Q Tokens}}} & \text{\textbf{\makecell{\# Avg. \\ Ans Tokens}}} & \text{\textbf{\makecell{\# Avg. Doc Tokens \\ (Top-5)}}} \\
\midrule
\multirow{3}{*}{\textbf{\makecell{New world \\ knowledge}}}
 & \makecell{\textbf{RealTimeQA} \\ \citep{kasai2023realtime}}   & 397                & 188                         & 19.0                   & 3.1                      & 216.7                            \\
& \makecell{\textbf{FreshQA} \\ \citep{vu2023freshllms}}    & 127                & 54                          & 13.8                   & 3.9                      & 227.5      
\\
\midrule
\multirow{4}{*}{\textbf{\makecell{Long-tail \\ knowledge}}}
& \makecell{\textbf{ToolQA} \\ \citep{zhuang2023toolqa}}      & 100                & 75                          & 21.7                   & 3.5                      & 425.3                            \\
& \makecell{\textbf{PopQA} \\ \citep{mallen-etal-2023-trust}}     & 1,399               & 659                         & 8.8                    & 4.0                      & 540.1                            \\
& \makecell{\textbf{TriviaQA} \\ \citep{joshi-etal-2017-triviaqa}}    & 7,313               & 295                        & 17.3                   & 5.9                      & 703.3                           \\
\midrule
\textbf{Total/Average} &  \texttt{RetrievalQA}  & 9,336               & 1,271                        & 13.2                   & 4.3                      & 510.1    \\
\bottomrule
\end{tabular}
\caption{
Data statistics of \texttt{RetrievalQA} (questions need retrieval). 
\textbf{\# Avg. Q, Ans, Doc Tokens} means the average number of tokens of questions, answers, and top-5 retrieved documents, respectively.
We use the \texttt{tiktoken} python library to calculate the number of tokens.
}
\label{tab_data_statistics}
\end{table*}
\endgroup

\subsection{Model Details}
\label{append_model}

We evaluate strong instruction-tuned models with a varying scale of model size: TinyLlama (1.1B) \citep{zhang2024tinyllama}, 
Phi-2\footnote{We acknowledge that Phi-2 has not been instruction fine-tuned; however, we find it performs decently well in understanding instructions.}
(2.7B) \citep{gunasekar2023textbooks, li2023textbooks}, 
Llama-2 (7B) \citep{touvron2023llama}, GPT-3.5 \citep{OpenAI_chatgpt2022}, and GPT-4 \citep{openai2023gpt4}.
We also use Self-RAG (7B, \citealp{asai2023selfrag}), which is fine-tuned based on Llama-2 7B using instruction-following corpora with interleaving passages and reflection tokens.
We download the models from HuggingFace\footnote{\url{https://huggingface.co/models}}. 
The model details, including downloading URLs, model size, and release date, can be found in Table \ref{tab_model_list}.

\begingroup
\renewcommand{\arraystretch}{1}
\begin{table*}[ht]
\small
\centering
\begin{NiceTabular}{lCc}[]
\toprule
  \B \text{Model Name} &  \text{\B Model Size} & \text{\B Release} \\ 
  \midrule
  \texttt{\href{https://huggingface.co/TinyLlama/TinyLlama-1.1B-Chat-v1.0}{TinyLlama/TinyLlama-1.1B-Chat-v1.0}} & 1.1\text{B} & Dec 2023 \\ 
  \texttt{\href{https://huggingface.co/microsoft/phi-2}{microsoft/phi-2}} & 2.7\text{B} & Dec 2023 \\
  \texttt{\href{https://huggingface.co/meta-llama/Llama-2-7b-chat-hf}{meta-llama/Llama-2-7b-chat-hf}} & 7\text{B} & Jul 2023  \\ 
  \texttt{\href{https://huggingface.co/selfrag/selfrag_llama2_7b}{selfrag/selfrag\_llama2\_7b}} & 7\text{B} & Oct 2023 \\ 
  \texttt{\href{https://platform.openai.com/docs/models/gpt-3-5-turbo}{gpt-3.5-turbo}} & \text{--} & Nov 2022 \\ 
  \texttt{\href{https://platform.openai.com/docs/models/gpt-4-and-gpt-4-turbo}{gpt-4-turbo-preview}} & \text{--} & Mar 2023 \\
\bottomrule
\end{NiceTabular}
\caption{Model used in the experiments.}
\label{tab_model_list}
\end{table*}
\endgroup

\begingroup
\renewcommand{\arraystretch}{1.2}
\begin{table}[ht]
\small
\centering
\begin{tabular}{lC}
\toprule
\textbf{Parameters}  & \textbf{Values} \\ 
  \midrule
\texttt{temperature} & 0.0  \\ 
\texttt{top\_p} & 1.0  \\
\texttt{max\_tokens} & 100  \\ 
Retrieved docs & \text{top-}5  \\ 
\makecell{Threshold \\ (Self-RAG)} & [\texttt{None}, 0.25, 0.5, 0.75]   \\ 
\makecell{\# demonstrations \\ (TA-ARE)} & 4   \\ 
Eval metric & \text{match/retrieval accuracy} \\
\bottomrule
\end{tabular}
\caption{Implementation hyperparameters.}
\label{tab_hyperparameters}
\end{table}
\endgroup

\subsection{Implementation Details}
\label{append_implementation_details}

% \paragraph{Implementation details.}
For fair comparisons, we use the same setting following Self-RAG for all experiments. The detailed hyperparameters are summarized in Table \ref{tab_hyperparameters}.

For Self-RAG, we set the retrieval threshold $t = [0.25, 0.5, 0.75, \texttt{None}]$. Lower thresholds encourage more frequent retrieval, while \texttt{None} means the model itself decides when to retrieve by generating the specific \texttt{[Retrieval]} token.
Since the quality of the retrieved documents is not the focus of this paper, we use the off-the-shelf Contriever \citep{izacard2022unsupervised} and author-provided top-5 documents extracted from Wikipedia where possible for long-tail knowledge questions. 
For questions from ToolQA, we use the author-provided vector database for retrieval of synthesized agendas.
Otherwise, we use top-5 documents returned by Google search\footnote{We use \href{https://serpapi.com/}{SerpApi} for Google search.} for new world knowledge questions.
To reduce API costs, for GPT-4, we randomly select 50 data instances from each source for evaluation, resulting in 250 questions.
We ask LLMs to respond \texttt{"I don't know"} if they cannot answer the question.

For instruction-tuned LMs, we use the official system prompt or instruction format used during training if publicly available.
We use vLLM \citep{10.1145/3600006.3613165} for accelerated inference.

\subsection{Sanity Check: Baselines Without Retrieval}
\label{append_no_retrieval}

We perform a sanity check on our \texttt{RetrievalQA} (questions need retrieval) using a simple QA template (Fig.\ref{fig_prompt_no_retrieval}) without retrieval.
As shown in Table \ref{tab_no_retrieval} and Fig.\ref{fig_sankey_gpt35}, all models achieve very poor match and F1 scores on \texttt{RetrievalQA}, indicating that it is extremely hard for models to answer the questions without consulting external resources.

We notice that TinyLlama, Phi-2, and Self-RAG have slightly better performance than larger models. 
Considering that these models were trained recently (as shown in Table \ref{tab_model_list}),
they might have learned some new knowledge and can answer some questions correctly.
Additionally, we conducted human checking on the questions answered correctly and found that some questions were mismarked due to multiple possible ground truths.
For example, for the question: \texttt{"Where will NeurIPS be located this year (2024)?"}, the model answers: \texttt{"NeurIPS will be held in Montreal, Canada."}, and the ground truth is an array of \texttt{["Vancouver, Canada", "Vancouver", "Canada"]}. Since the model prediction contains \texttt{Canada}, this answer was marked correct.
However, 
% \textbf{LLMs still do not \textit{truly} know the answer without retrieval}.
\textbf{LLMs themselves still do not \textit{truly} know the answer}.
The outdated knowledge stored in their parameters makes them hallucinate.
% Second, these models were trained recently (as shown in Table \ref{tab_model_list}) and might have learned some new knowledge; therefore, they might answer some questions correctly.
Since these questions only take a tiny portion of the entire dataset (as an example shown in Fig.\ref{fig_sankey_gpt35}, the  \colorbox{lightpink!100}{tiny red line}  from \texttt{Retrieval=No} to \texttt{Prediction=Correct}), 
and early-trained models such as Llama-2 and GPT-3/4 perform worse, we still keep them in our dataset.

\begin{table}[ht]
\centering
\small
\begin{NiceTabular}{lCC}
\toprule
\textbf{Model}  & \textbf{Match} & \textbf{F1} \\ \midrule
TinyLlama (1.1B)   & 4.2   & 1.3   \\
Phi-2 (2.7B)     & 7.2   & 3.9    \\
Llama-2 (7B) & 2.0   & 0.7     \\
Self-RAG (7B) & 6.0   &  1.5    \\
GPT-3.5   & 1.2  & 1.0   \\
GPT-4*   & 2.4   &  2.3  \\
\bottomrule
\end{NiceTabular}
\caption{Match and F1 scores of models on \texttt{RetrievalQA} (1,271) \textbf{without} retrieval. * indicates that we evaluate GPT-4 using 250 examples to reduce API costs.}
\label{tab_no_retrieval}
\end{table}

\begin{table}[ht]
\centering
\small
\begin{NiceTabular}{lC}
\toprule
\textbf{Model}  & \textbf{Match} \\ \midrule
TinyLlama (1.1B)   & 88.1     \\
Phi-2 (2.7B)     & 87.7     \\
Llama-2 (7B) & 89.8      \\
Self-RAG (7B) & 88.2       \\
GPT-3.5   & 91.1    \\
GPT-4*   & 88.4    \\
\bottomrule
\end{NiceTabular}
\caption{Match scores of models on 1,517 questions that do not need retrieval.}
\label{tab_no_retrieval_1514}
\end{table}

%%%%%%%%%%%%%%%%%%%%%% fewshot_no_time %%%%%%%%%%%%%%%%%%%%%%
\begin{table}[ht]
\small
\centering
\begin{tabular}{lCC}
\toprule
\multirow{2}[3]{*}{\B Baselines}   & \multicolumn{2}{c}{\textbf{Adaptive Retrieval}} \\ 
\cmidrule(lr){2-3} & \text{Retrieval} & \text{Match}  \\
\midrule
TinyLlama (1.1B) & 90.9 \ \text{\footnotesize{(\G{+51.8})}} & 27.5 
 \ \text{\footnotesize{(\G{+12.8})}}  \\
Phi-2 (2.7B)     & 88.7 \ \text{\footnotesize{(\R{-5.4})}} & 33.8 \ \text{\footnotesize{(\R{-1.2})}}   \\
Llama-2 (7B)     & 47.0 \ \text{\footnotesize{(\R{-33.3})}} & 16.7 \ \text{\footnotesize{(\R{-9.4})}}   \\
% Self-RAG (7B) $_{\texttt{t=None}}$ & 0.4  & 6.0 & 31.9 \\
GPT-3.5          & 87.6 \ \text{\footnotesize{(\G{+38.3})}} & 36.2 \ \text{\footnotesize{(\G{+15.4})}}      \\
GPT-4            & 86.0 \ \text{\footnotesize{(\G{+18.4})}} & 46.0 \ \text{\footnotesize{(\G{+8.4})}}  \\  

\midrule
Average gain & \G{\text{+}14.0} & \G{\text{+}5.2} \\

\bottomrule
\end{tabular}
\caption{
Ablation: our \textbf{TA-ARE} without the current date.
% \textbf{ICL without time} prompting retrieval and match accuracy on \texttt{RetrievalQA}.
\R{(-red)} means performance losses compared to \textbf{Vanilla} prompting in Table \ref{tab_vanilla_results}.
}
\label{tab_fewshot_no_time_results}
\end{table}
%%%%%%%%%%%%%%%%%%%%%% fewshot_no_time %%%%%%%%%%%%%%%%%%%%%%

%%%%%%%%%%%%%%%%%%%%%% no_fewshot_time %%%%%%%%%%%%%%%%%%%%%%
\begin{table}[ht]
\small
\centering
\begin{tabular}{lCC}
\toprule
\multirow{2}[3]{*}{\B Baselines}   & \multicolumn{2}{c}{\textbf{Adaptive Retrieval}} \\ 
\cmidrule(lr){2-3} & \text{Retrieval} & \text{Match}  \\
\midrule
TinyLlama (1.1B) & 73.8 \ \text{\footnotesize{(\G{+34.7})}} & 23.1 
 \ \text{\footnotesize{(\G{+8.4})}}  \\
Phi-2 (2.7B)     & 89.5 \ \text{\footnotesize{(\R{-4.6})}} & 32.2 \ \text{\footnotesize{(\R{-2.8})}}   \\
Llama-2 (7B)     & 90.0 \ \text{\footnotesize{(\G{+9.7})}} & 31.4 \ \text{\footnotesize{(\G{+5.3})}}   \\
% Self-RAG (7B) $_{\texttt{t=None}}$ & 0.4  & 6.0 & 31.9 \\
GPT-3.5          & 36.7 \ \text{\footnotesize{(\R{-12.6})}} & 18.9 \ \text{\footnotesize{(\R{-1.9})}}      \\
GPT-4            & 70.0 \ \text{\footnotesize{(\G{+2.4})}} & 39.6 \ \text{\footnotesize{(\G{+2.0})}}  \\  

\midrule
Average gain & \G{\text{+}5.9} & \G{\text{+}2.2} \\

\bottomrule
\end{tabular}
\caption{
Ablation: our \textbf{TA-ARE} without demonstration examples.
% \textbf{ICL without time} prompting retrieval and match accuracy on \texttt{RetrievalQA}.
% \textbf{Only time} prompting retrieval and match accuracy on \texttt{RetrievalQA}.
% \R{(-red)} means performance losses compared to \textbf{Direct} prompting.
}
\label{tab_no_fewshot_time_results}
\end{table}
%%%%%%%%%%%%%%%%%%%%%% no_fewshot_time %%%%%%%%%%%%%%%%%%%%%%

We also run the sanity check on the 1,514 questions that do not need retrieval. As shown in Table \ref{tab_no_retrieval_1514}, even strong models like GPT-3.5 and GPT-4 can not reach 100\% match accuracy using their parametric knowledge.

\begin{table*}[t]
\small
% \footnotesize
% \scriptsize
\centering
\begin{tabular}{lCCCCCCC}
\toprule
\multirow{2}[3]{*}{\B Baselines (2,785)} & \textbf{\makecell{No Retrieval}}  & \multicolumn{5}{c}{\textbf{Adaptive Retrieval}} & \textbf{\makecell{Always \\ Retrieval}} \\ 

\cmidrule(lr){3-7} & \text{Match} & \text{Match} & \text{Retrieval Acc} & \text{Precision} & \text{Recall} & \text{F1} & \text{Match}                  \\
\midrule

& \multicolumn{7}{c}{\textbf{\textit{Calibration-based}}} \\

Self-RAG (7B) & & & \\
\quad $t=0.25$ & \multirow{4}{*}{$50.7$} & 	64.3 & 45.6 & 50.0 & 22.8 & 31.3 & \multirow{4}{*}{$64.3$} \\
\quad $t=0.5$  &  & 53.2  &  53.6 &  51.2 & 51.7 & 51.5	&  \\
\quad $t=0.75$ &  & 50.7  &  54.4 &  50.0 & 27.2 & 35.2	& \\
\quad $t=\texttt{None}$ & & 49.3  & 54.5 & 50.1  & 62.9 & 55.8	& \\

\midrule

% \cdashline{1-4}

 & \multicolumn{7}{c}{\textbf{\textit{Model-based}}} \\
\cellcolor{lightgray!25} \textbf{\textit{Vanilla}} \ \cref{sec_preliminary}  & \cellcolor{lightgray!25} & \cellcolor{lightgray!25} & \cellcolor{lightgray!25} & \cellcolor{lightgray!25} & \cellcolor{lightgray!25} & \cellcolor{lightgray!25} & \cellcolor{lightgray!25} \\

TinyLlama (1.1B) & 49.8 & 54.4 & 49.3 & 48.5 &  48.4 & 48.5	& 59.9 \\
Phi-2 (2.7B)     & 	51.0 & 64.9 & 48.0 & 51.7 &  55.8 & 53.7	& 65.7   \\
Llama-2 (7B)     & 49.7 & 60.4 & 44.3 & 47.2 &  45.0 & 46.1		& 65.8   \\
% Self-RAG (7B) $_{\texttt{t=None}}$ & 0.4  & 6.0 & 31.9 \\
GPT-3.5          & 50.1 & 58.7 & 61.3 & 60.3 & 60.9& 60.6		& 65.7  \\
GPT-4*          & 45.4 & 64.4 & 76.0 & 	76.0 & 76.8 & 76.4		& 64.2 \\

% \midrule
% \cdashedline{1-4}
% \cmidrule(l){1-4}
% \midrule[dashed]
\hdashline

%  & \multicolumn{3}{c}{\textbf{\textit{Model-based}}} \\
\cellcolor{lightgray!25} \textbf{\textit{Ours TA-ARE}} \ \cref{sec_icare}  & \cellcolor{lightgray!25} & \cellcolor{lightgray!25} & \cellcolor{lightgray!25} & \cellcolor{lightgray!25} & \cellcolor{lightgray!25} & \cellcolor{lightgray!25} & \cellcolor{lightgray!25}\\

TinyLlama (1.1B) & 49.8 & 56.1 & 44.7 & 45.4 & 45.3 & 45.4& 59.9 \\
Phi-2 (2.7B)     & 51.0 & 65.6 & 54.1 & 57.4 & 66.8 & 61.8 & 65.7 \\
Llama-2 (7B)     & 49.7 & 63.3 & 44.3 & 47.6 & 44.2 & 45.8	& 65.8 \\
% Self-RAG (7B) $_{\texttt{t=None}}$ & 0.4  & 6.0 & 31.9 \\
GPT-3.5          & 50.1 & 65.3 & 67.1 & 68.6 & 70.6 & 69.6 & 65.7  \\
GPT-4*           & 45.4 & 67.6 & 76.6 & 76.6 & 77.1 & 76.8 & 64.2 \\  

% \midrule
Average gain &  \text{--} & \G{\text{+}3.0} & \G{\text{+}1.6} & \G{\text{+}2.4} &  \G{\text{+}3.4} & \G{\text{+}2.8} &  \text{--} \\

\bottomrule
\end{tabular}
\caption{
% \footnotesize
% \textbf{Vanilla} prompting 
Retrieval and match accuracy on \texttt{RetrievalQA} (overall). 
% For Self-RAG, we set the retrieval threshold $t = [0.25, 0.5, 0.75, \texttt{None}]$. \texttt{None} means the model itself decides when to retrieve.
* indicates using 500 examples for testing to reduce API costs.
% Best scores in \textbf{Bold} and second best in \underline{underline}.
}
\label{tab_overall_results}
\end{table*}

\subsection{Overall Results}
\label{appendix_overall_results}

In Table \ref{tab_vanilla_results}, we only evaluate 1,271 questions that need retrieval. In this section, we evaluate the entire 2,785 data, with 1,271 labelled as required retrieval and 1,514 labelled as do not require retrieval. Besides retrieval accuracy, we also report retrieval macro precision, recall, and F1.
Table \ref{tab_overall_results} shows the overall results. Using questions that do not need retrieval and questions that need retrieval, we have a comprehensive evaluation of ARAG methods.

%%%%%%%%%%%%%%%%%%%%%% data example %%%%%%%%%%%%%%%%%%%%%%
\begingroup
\begin{table*}[ht]
% \small
\footnotesize
% \scriptsize
\centering
\begin{tabular}{cccl}
\toprule
\textbf{Category} & \textbf{Data Source} & \textbf{Question} & \textbf{Answer}  \\
\midrule
\multirow{3}{*}{\textbf{\makecell{New world \\ knowledge}}}
 & \makecell{\textbf{RealTimeQA} \\ \citep{kasai2023realtime}}   &  \makecell{Which 2024 Republican presidential contender \\ announced that he is ending his campaign?} & Former Texas Rep. Will Hurd  \\
& \makecell{\textbf{FreshQA} \\ \citep{vu2023freshllms}}   & \makecell{What is the latest highest-grossing \\ movie of the week at the Box office?} & Mean Girls    
\\
\midrule
\multirow{5}{*}{\textbf{\makecell{Long-tail \\ knowledge}}}
& \makecell{\textbf{ToolQA} \\ \citep{zhuang2023toolqa}}      & \makecell{What time did Grace attend Broadway \\ Show on 2022/02/17?} & 8:00 PM  \\
& \makecell{\textbf{PopQA} \\ \citep{mallen-etal-2023-trust}}     & What is Henry Feilden's occupation? &  politician \\
& \makecell{\textbf{TriviaQA} \\ \citep{joshi-etal-2017-triviaqa}}    & \makecell{Which bird, that breeds in northern Europe \\ in pine and beech forests, has a chestnut \\ brown back, grey head, dark tail, buff \\ breast and a striped black throat?} & fieldfare \\

\bottomrule
\end{tabular}
\caption{
Data examples of \texttt{RetrievalQA} (questions need external retrieval). 
}
\label{tab_data_instances}
\end{table*}
\endgroup
%%%%%%%%%%%%%%%%%%%%%% data example %%%%%%%%%%%%%%%%%%%%%%

\subsection{Prompt Templates}
\label{appendix_prompt}

We present the prompt templates used in the experiments, as shown in Fig.\ref{fig_prompt_vanilla}, 
Fig.\ref{fig_prompt_icare}, Fig.\ref{fig_prompt_no_retrieval}, Fig.\ref{fig_prompt_retrieval}. 
% For instruction-tuned LLMs, we use the official system prompt or instruction format used during training if publicly available.

\begin{figure*}
    \begin{tikzpicture}
    \node [draw, rounded corners,
         text width=\linewidth-24pt,    % <---
         align=flush center, 
         inner sep=12 pt,
         fill=lightgray,
    ]%
    {
    \begin{minipage}{\linewidth}
      Given a question, determine whether you need to retrieve external resources, such as real-time search engines, Wikipedia, or databases, to answer the question correctly. Only answer "[Yes]" or "[No]".\\
    
      Question: \{\texttt{question}\} \\
      Answer:      
    \end{minipage}
     };
\end{tikzpicture}
\caption{
% Zero-shot instruction prompt template for explicitly asking whether retrieval is needed.
\textbf{Vanilla} prompt template for adaptive retrieval  (\cref{sec_adaptive_rag_method}).
}
\label{fig_prompt_vanilla}
\end{figure*}

\begin{figure*}
    \begin{tikzpicture}
    \node [draw, rounded corners,
         text width=\linewidth-24pt,    % <---
         align=flush center, 
         inner sep=12 pt,
         fill=lightgray,
    ]%
    {
    \begin{minipage}{\linewidth}
      Today is \{\texttt{datetime.today()}\}. Given a question, determine whether you need to retrieve external resources, such as real-time search engines, Wikipedia, or databases, to answer the question correctly. Only answer "[Yes]" or "[No]".\\

      Here are some examples:\\
      \{\texttt{demonstration examples}\}\\
    
      Question: \{\texttt{question}\} \\
      Answer:      
    \end{minipage}
     };
\end{tikzpicture}
\caption{
Ours \textbf{TA-ARE} prompt template for adaptive retrieval  (\cref{sec_icare}).
}
\label{fig_prompt_icare}
\end{figure*}

\begin{figure*}
    \begin{tikzpicture}
    \node [draw, rounded corners,
         text width=\linewidth-24pt,    % <---
         align=flush center, 
         inner sep=12 pt,
         fill=lightgray,
    ]%
    {
    \begin{minipage}{\linewidth}
      Please use your own knowledge to answer the questions. Only include the answer in your response and try to be concise. If you do not know the answer, just say "I don't know".\\
    
      Question: \{\texttt{question}\} \\
      Answer:      
    \end{minipage}
     };
\end{tikzpicture}
\caption{
Instruction prompt template for QA without retrieval.
}
\label{fig_prompt_no_retrieval}
\end{figure*}

\begin{figure*}
    \begin{tikzpicture}
    \node [draw, rounded corners,
         text width=\linewidth-24pt,    % <---
         align=flush center, 
         inner sep=12 pt,
         fill=lightgray,
    ]%
    {
    \begin{minipage}{\linewidth}
      Please answer the question based on the provided context. Only include the answer in your response and try to be concise. If you do not know the answer, just say "I don't know".\\
      
      Paragraph: \\
      \{\texttt{retrieved documents}\} \\
    
      Question: \{\texttt{question}\} \\
      Answer:      
    \end{minipage}
     };
\end{tikzpicture}
\caption{
Instruction prompt template for QA with retrieved documents.
}
\label{fig_prompt_retrieval}
\end{figure*}

\end{document}